\documentclass[sigconf]{acmart}
\AtBeginDocument{%
  }

\setcopyright{none}
\copyrightyear{2018}
\acmYear{2018}
\acmDOI{XXXXXXX.XXXXXXX}
\renewcommand{\shortauthors}{}
\AtBeginDocument{\pagestyle{empty}}
\acmConference[Conference acronym 'XX]{Make sure to enter the correct
  conference title from your rights confirmation email}{June 03--05,
  2018}{Woodstock, NY}
\acmISBN{978-1-4503-XXXX-X/2018/06}
\usepackage{amsmath}
\usepackage{wrapfig}
\usepackage{algorithm}
\usepackage{algorithmic}
\usepackage{fancyvrb}
\usepackage{varwidth}
\usepackage{float}
\usepackage{framed}
\usepackage{verbatimbox}
\usepackage{multirow}
\usepackage{pgfplots}
\pgfplotsset{compat=1.18}
\usepackage{dblfloatfix}
\usepackage{placeins}

\begin{document}

\title{Node-as-Agent: Graph Agentic Network}

\author{Minghao Guo}
\affiliation{%
  \institution{Rutgers University}
  \city{New Brunswick}
  \state{NJ}
  \country{USA}}
\email{minghao.guo@rutgers.edu}

\author{Xi Zhu}
\affiliation{%
  \institution{Rutgers University}
  \city{New Brunswick}
  \state{NJ}
  \country{USA}}
\email{xi.zhu@rutgers.edu}

\author{Qingyue Jiao}
\affiliation{%
  \institution{University of Notre Dame}
  \streetaddress{Holy Cross Dr.}
  \city{Notre Dame}
  \state{IN}
  \postcode{46556}
  \country{USA}}
\email{jiao@nd.edu}

\author{Xiujin Liu}
\affiliation{%
  \institution{University of Michigan--Ann Arbor}
  \city{Ann Arbor}
  \state{MI}
  \country{USA}}
\email{jeanliu@umich.edu}

\author{Haochen Xue}
\affiliation{%
  \institution{University of Liverpool}
  \city{Liverpool}
  \country{UK}}
\email{hicaca945@gmail.com}

\author{Chong Zhang}
\affiliation{%
  \institution{University of Liverpool}
  \city{Liverpool}
  \country{UK}}
\email{C.Zhang118@liverpool.ac.uk}

\author{Shuhang Lin}
\affiliation{%
  \institution{Rutgers University}
  \city{New Brunswick}
  \state{NJ}
  \country{USA}}
\email{shuhang.lin@rutgers.edu}

\author{Jingyuan Huang}
\affiliation{%
  \institution{Rutgers University}
  \city{New Brunswick}
  \state{NJ}
  \country{USA}}
\email{chy.huang@rutgers.edu}

\author{Ziyi Ye}
\affiliation{%
  \institution{Fudan University}
  \city{Shanghai}
  \country{China}}
\email{zyye@fudan.edu.cn}

\author{Yongfeng Zhang}
\affiliation{%
  \institution{Rutgers University}
  \city{New Brunswick}
  \state{NJ}
  \country{USA}}
\email{yongfeng.zhang@rutgers.edu}

\renewcommand{\shortauthors}{Minghao Guo et al.}

\begin{abstract}
Graph Neural Networks~(GNNs) have achieved remarkable success in graph-based learning by propagating information among neighbor nodes through predefined aggregation mechanisms. 
However, such fixed schemes often suffer from two key limitations. 
On the one hand, it cannot handle the varying node informativeness—some nodes are rich in information, while others remain sparse. 
On the other hand, predefined message passing primarily leverages local structural similarity while ignoring global semantic relationships across the graph, limiting the model’s ability to capture distant but relevant information.
To address these limitations, we propose the Retrieval-augmented Graph Agentic Network (ReaGAN), an agent-based framework that addresses these limitations by empowering each node with autonomous, individual node-level decision-making. 
Each node is treated as an agent that independently plans its next action based on its internal memory, enabling node-level planning and adaptive message propagation. 
Furthermore, we integrate retrieval-augmented generation (RAG) as a tool for agents to dynamically build global relationships by retrieving semantically relevant content from across the graph.
Extensive experiments demonstrate that ReaGAN achieves competitive performance under few-shot in-context settings, using only a frozen LLM backbone without fine-tuning. 
These results highlight the potential of agentic planning and integrated local-global retrieval for advancing graph machine learning.
\end{abstract}




\maketitle

\section{Introduction}
\label{sec:introduction}

Graph Machine Learning (GML) has achieved remarkable success over the past years, with Graph Neural Networks (GNNs) such as GCN~\citep{kipf2017semi}, GAT~\citep{velivckovic2017graph}, and GraphSAGE~\citep{hamilton2017inductive} becoming the de facto standards for representation learning over graph-structured data. These models operate through a static, globally synchronized message-passing framework, where in each layer, every node aggregates information from its neighbors using a predefined aggregation and update rule parameterized by shared weights across the graph.

\begin{figure}[H]
    \centering
    \includegraphics[width=1.0\linewidth]{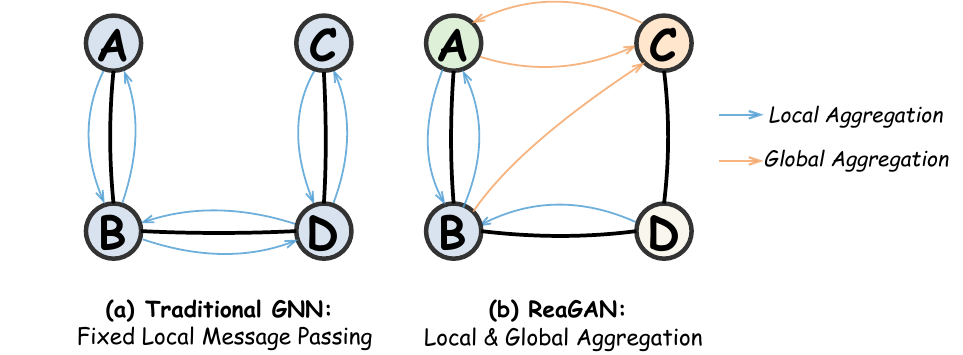}
    \caption{Message passing in traditional pre-defined way vs. ReaGAN's method.
    (a) All nodes perform local aggregation only.
    (b) Nodes aggregate in different ways. A aggregates both locally and globally; B performs local-only aggregation; C performs global-only aggregation; D performs no operation.
    }
    \label{fig:message_passing}
\end{figure}

While effective in many scenarios, this paradigm suffers from a fundamental limitation: it treats all nodes uniformly, regardless of their varying local context or inherent semantics~\citep{finkelshtein2023cognn, martinkus2023agent}. 
The aggregation-update process is homogeneous: each node follows the same layer-wise rule, and no node has the capacity for individualized decision-making.
However, graphs often contain a small subset of nodes that are rich in semantic content or structurally well-positioned, while others are sparsely connected, noisy, or contextually ambiguous. Thus, applying identical message-passing rules to all nodes in such settings is problematic: highly informative nodes may be overwhelmed by irrelevant inputs, while uninformative nodes fail to gather the support they need. Worse still, this homogeneous propagation can amplify noise and redundancy, thereby degrading overall representation quality. This observation motivates \textbf{\textit{Challenge 1: Node-Level Autonomy}} — \textit{Can we endow each node with the capacity to autonomously plan its own message-passing behavior based on its internal state and local context, rather than relying on globally shared rules?} 

In addition, most existing GNNs operate primarily based on local structural similarity, implicitly assuming that neighboring nodes share informative and task-relevant features. However, this assumption often breaks down in real-world graphs where semantically related nodes may be structurally distant, which is a pattern especially common in open-domain or heterogeneous networks~\citep{zhu2024llm, li2024incontext}. As a result, traditional GNNs struggle to capture global semantic dependencies, limiting their ability to perform long-range reasoning or generalize beyond local structure. This limitation is especially pronounced in sparsely connected or noisy regions, where the local neighborhoods offer limited predictive value~\citep{li2024incontext, gao2026beyond}. In such settings, retrieving semantically aligned but distant nodes becomes essential for enriching context and improving node representations. This raises  \textit{\textbf{Challenge 2: Local-Global Complementarity}} — \textit{Can we combine local-structure neighbors with global-semantic neighbors to enable a more comprehensive context-aware message passing?}


To address the two key challenges, we propose a new perspective on graph learning by rethinking the fundamental computational unit: \textbf{\textit{we treat each node as an autonomous agent that decides its own aggregation actions.}} Unlike GNNs that passively aggregate messages through static rules, each node acts as an agent that actively determines its action at each layer based on internal states and contextual signals.
Inspired by general agent-based systems and LLM-based simulation environments~\citep{russell2010artificial, wang2023surveyagent, xu2025amemagenticmemoryllm, sunvisual, xu2026ael, du2025twinvoice, guo2026individual, li2025know, nie2024facttest}, our agentic framework endows each node with four core components:  \textit{\textbf{Planning}}, which decides the next operations based on its current objective and context by a frozen LLM; \textit{\textbf{Actions}}, which execute local-global aggregation strategies to interact with neighbors~\citep{yao2022react, shinn2023reflexion}; \textit{\textbf{Memory}}, which stores the node's cumulative textual features along with selected neighbor features; and \textit{\textbf{Tool-Use}}, which calls external functions like retrieval-augmented generation to enhance the feature representation.
In this work, we unify these components into the \textbf{Retrieval-augmented Graph Agentic Network (ReaGAN)}, which enables each node to make individualized, adaptive, and semantically informed decisions. This self-decision-making design directly addresses Challenge 1, while the local-global aggregation strategy addresses Challenge 2 as the following paragraph shows.

Beyond Local Aggregation, which enables nodes to exchange messages with directly connected neighbors, we integrate retrieval-augmented generation (RAG)~\citep{lewis2020retrieval,10.1145/3726302.3729957} as an external tool that empowers nodes to access semantically relevant but structurally distant information, where the entire graph is viewed as a searchable database. Each node then integrates both local structural signals and globally retrieved semantic content into its memory, forming a richer contextual representation for subsequent planning and interaction. Equipped with these capabilities, each node engages in an agentic workflow. Based on its current state and memory, the node prompts a frozen LLM to generate a context-aware plan, selects and executes the appropriate actions, such as local or global aggregation, optionally invokes external tools like RAG, and subsequently updates its internal memory with the acquired information.

We pinpoint the key differences between traditional message passing and our agentic workflow. As shown in Figure ~\ref{fig:message_passing}(a), standard GNNs enforce a uniform propagation rule, where all nodes communicate only with immediate neighbors. In contrast, Figure ~\ref{fig:message_passing}(b) illustrates ReaGAN’s mechanism,  in which each node autonomously selects its own actions and possibly access semantically similar but structurally distant neighbors. For example, node A performs both local and global aggregation; node B conducts only local aggregation, resembling classical GNN behavior; node C executes only global aggregation; and node D chooses to remain inactive. This demonstrates how ReaGAN supports diverse and asynchronous strategies at the node level, offering node-level flexibility.

Crucially, our entire agentic framework is powered by a frozen, off-the-shelf Large Language Model (LLM) that serves as the core reasoning engine for each node agent. This design choice is fundamental to our contribution and carries significant implications. By leveraging a pre-trained and frozen LLM, we completely bypass the need for costly, gradient-based training and task-specific fine-tuning, which are hallmarks of traditional GNNs. This approach positions ReaGAN not merely as a new architecture, but as a more resource-efficient and flexible ``plug-and-play'' paradigm for graph learning. It allows us to directly tap into the powerful, generalized reasoning capabilities of the LLM, enabling strong performance in few-shot scenarios without learning any trainable parameters for the graph task at hand.

\noindent We summarize our main contributions as follows:
\begin{itemize}
    \item We introduce ReaGAN, an agentic graph learning framework that models each node as an autonomous agent equipped with planning, memory, action, and tool-use capabilities, thereby moving beyond the static, rule-based message passing inherent in traditional GNNs.

    \item We introduce a hybrid aggregation mechanism that integrates local structural and global semantic information via retrieval-augmented generation, allowing nodes to dynamically access semantically relevant but structurally distant context.

    \item We conduct extensive experiments to demonstrate that ReaGAN achieves competitive performance compared to existing traditional supervised methods, even when using a frozen LLM without fine-tuning.
\end{itemize} 

\section{Method}

\subsection{Node as Agent}  
In ReaGAN, each node is treated as an autonomous agent capable of perceiving its own state and neighborhood context, planning its next steps, executing context-aware actions, utilizing external tools, and updating its internal memory. This perspective departs from traditional synchronized message passing, instead enabling fully individualized, asynchronous, and adaptive behavior. Moreover, each node can aggregate from local structural neighbors, and retrieve from global semantic neighbors. By doing so, ReaGAN naturally supports both \textit{node-level personalization and autonomy} (addressing \textit{Challenge 1}) and \textit{joint local-global information integration} (addressing \textit{Challenge 2}).


\subsection{Agentic Formulation}
Let $\mathcal{G} = (\mathcal{V}, \mathcal{E})$ be an attributed graph, where each node $v \in \mathcal{V}$ is associated with a text feature $t_v$ and an optional label $y_v \in \mathcal{Y}$. The goal is to predict labels $\hat{y}_v$ for all unlabeled nodes.
Each node is treated as an agent equipped with memory $\mathcal{M}_v$, interacting with a frozen LLM through a multi-layered reasoning loop. At each layer, the node constructs a prompt from its previous-layer memory $\mathcal{M}_v^{(l-1)}$, queries the LLM for an action plan, and then updates its memory. After $L$ layers, the node queries the LLM once for prediction.

In summary, each node follows a layer-wise cycle of \textit{perception}, \textit{planning}, \textit{action execution}, and \textit{memory update}, independently deciding whether to gather local/global information, make a prediction, or take no action, and the label is predicted only at the final layer. This fully agentic workflow is detailed in Algorithm~\ref{alg:reasoning-loop}.

\begin{algorithm}[ht]
\caption{Layer-wise reasoning loop for agent node $v$ in ReaGAN}
\label{alg:reasoning-loop}
\begin{algorithmic}[1]
\REQUIRE Text-attributed graph $\mathcal{G}$, node $v$, frozen LLM, retrieval database $\mathcal{D}$
\STATE Initialize memory: $\mathcal{M}_v^{(0)} \leftarrow \{ t_v \}$
\STATE Initialize aggregated feature: $\tilde{t}_v^{(0)} \leftarrow t_v$
\FOR{layer $l = 1$ to $L$}
    \STATE $p_v^{(l)} \leftarrow \texttt{Prompt}_{\texttt{planning}}(\mathcal{M}_v^{(l-1)})$
    \STATE $a_v^{(l)} \leftarrow \texttt{LLM}(p_v^{(l)})$
    \FOR{action $a$ in $a_v^{(l)}$}
        \IF{$a = \texttt{LocalAggregation}$}
            \STATE $\tilde{t}_v^{(l)} \leftarrow \texttt{TextAgg}(\tilde{t}_v^{(l-1)}, \{\tilde{t}_u^{(l-1)} \mid u \in \mathcal{N}_{\text{local}}(v) \})$
            \STATE $\mathcal{E}_v^{(l)} \leftarrow \{ (\tilde{t}_u^{(l-1)}, y_u) \mid u \in \mathcal{N}_{\text{local}}(v), y_u \in \mathcal{Y} \}$
            \STATE $\mathcal{M}_v^{(l)} \leftarrow \mathcal{M}_v^{(l-1)} \cup \{ \tilde{t}_v^{(l)} \} \cup \mathcal{E}_v^{(l)}$
        \ELSIF{$a = \texttt{GlobalAggregation}$}
            \STATE $\mathcal{N}_{\text{global}}(v) \leftarrow \texttt{RAG}(\tilde{t}_v^{(l-1)}, \texttt{top} = K)$
            \STATE $\tilde{t}_v^{(l)} \leftarrow \texttt{TextAgg}(\tilde{t}_v^{(l-1)}, \{ \tilde{t}_u^{(l-1)} \mid u \in \mathcal{N}_{\text{global}}(v) \})$
            \STATE $\mathcal{E}_v^{(l)} \leftarrow \{ (\tilde{t}_u^{(l-1)}, y_u) \mid u \in \mathcal{N}_{\text{global}}(v), y_u \in \mathcal{Y} \}$
            \STATE $\mathcal{M}_v^{(l)} \leftarrow \mathcal{M}_v^{(l-1)} \cup \{ \tilde{t}_v^{(l)} \} \cup \mathcal{E}_v^{(l)}$
        \ELSIF{$a = \texttt{NoOp}$}
            \STATE $\mathcal{M}_v^{(l)} \leftarrow \mathcal{M}_v^{(l-1)}$ \COMMENT{No change to memory}
        \ENDIF
    \ENDFOR
\ENDFOR
\RETURN Predicted label $\hat{y}_v$ (if generated)
\end{algorithmic}
\end{algorithm}

\subsection{Planning: Prompting Frozen LLM}
\label{sec:planning}

\begin{figure*}[t]
    \centering
    \includegraphics[width=1.0\linewidth]{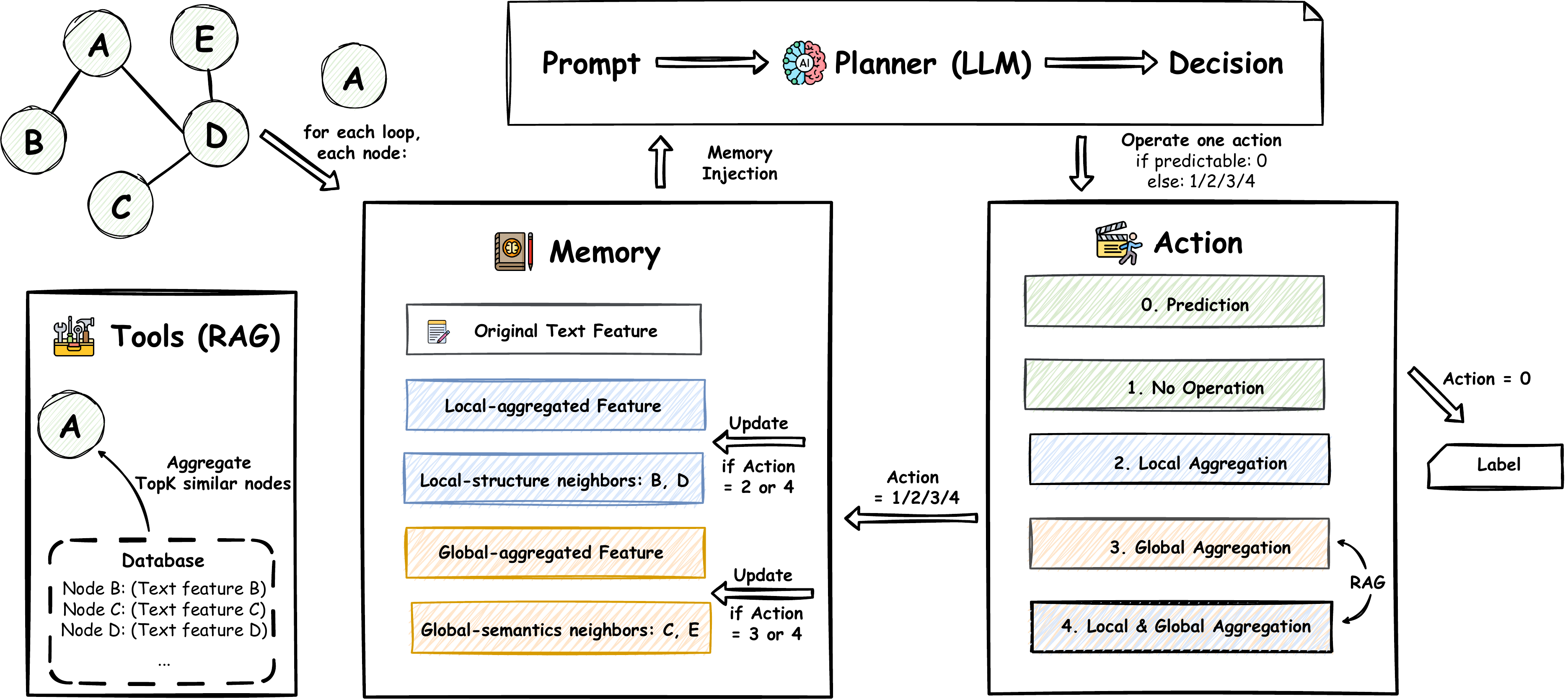}
    \caption{Overview of ReaGAN. Each node in ReaGAN is modeled as an agent equipped with four core modules: \textit{Memory}, \textit{Planning}, \textit{Tools}, and \textit{Action}. 
    The node stores its original information and receives local and global information into its memory, constructs a natural language prompt, and queries a frozen LLM for an action plan. 
   The planner outputs one action per layer (e.g., Local Aggregation, Global Aggregation), which are executed using available tools such as RAG. 
    The resulting outputs are written back into memory and may produce a predicted label. 
    This forms a closed loop of perception, planning, action, and memory refinement across layers.
    }
    \label{fig:gan-overview}

\end{figure*}

In ReaGAN, each node is equipped with a local planner that determines how to act at each layer. Rather than relying on a globally shared aggregation rule, we leverage a frozen LLM to enable in-context planning, where decisions are conditioned on the node’s own memory. At each layer $l$, node $v$ constructs a structured prompt based on its internal memory $\mathcal{M}_v^{(l)}$. The prompt is then sent to a frozen LLM (e.g., LLaMA~\citep{touvron2023llama}, Qwen~\citep{baizhou2024qwen}, or DeepSeek~\citep{deepseek2024open}), which returns an action plan:

\[
    a_v^{(l)} = \texttt{LLM}(\texttt{Prompt}_{\texttt{planning}}(\mathcal{M}_v^{(l-1)}))
\]

where the action $a_v^{(l)}$ is selected from a discrete space and described in detail in Section~\ref{sec:action}. The node then parses and executes the plan, updating its memory with newly acquired information. This process is repeated for $L$ layers. At the final layer, the node constructs a prediction-specific prompt and queries the LLM to produce a label:

\[
    \hat{y}_v = \texttt{LLM}(\texttt{Prompt}_{\texttt{predict}}(\mathcal{M}_v^{(L)}))
\]

which concludes the planning loop illustrated in Figure~\ref{fig:gan-overview}. To sum up, the planning process allows each node to reason independently and asynchronously, determining what to do and when to act—without any global synchronization. It forms the core mechanism that enables \textit{Challenge 1 (Node-Level Autonomy)} and \textit{Challenge 2 (Global Semantic Access)} to be addressed in a unified and decentralized manner.

\subsection{Action: Node-Level Decision Space}
\label{sec:action}

The purpose of operating actions is to aggregate information, both in local and global ways. The aggregation process includes enhancing the text feature and collecting neighbor shots. The details will be discussed in the following sections.

\subsubsection{Local Aggregation.}

We refer to directly connected nodes in the input graph as \textit{local structural neighbors}, and to semantically similar but unconnected nodes retrieved via RAG as \textit{global semantic neighbors}. When a node selects the \texttt{Local Aggregation} action, it gathers information from its local structural neighbors (e.g., 1-hop or 2-hop nodes). This action serves two primary purposes:

\begin{itemize}
    \item \textbf{Feature Enhancement.}  
    All neighbors, regardless of label availability, contribute to a contextual text aggregation process. The node summarizes the textual content of its local neighborhood—typically by generating a single aggregated text snippet $\tilde{t}_v^{(l)}$ that combines its own feature $t_v$ with those of its structural neighbors. This mirrors the feature aggregation in GNNs, where embeddings are iteratively enriched from local neighbors. The aggregated feature is defined as:
    \[
    \tilde{t}_v^{(l)} = \texttt{TextAgg} \left( \tilde{t}_v^{(l-1)}, \left\{ \tilde{t}_u^{(l-1)} \,\middle|\, u \in \mathcal{N}_{\text{local}}(v) \right\} \right)
    \]
    where $\texttt{TextAgg}(\cdot)$ denotes a natural language-level aggregation function such as concatenation or summarization. Here, $\tilde{t}_v^{(l-1)}$ is the node’s aggregated feature from the previous layer (initialized as $t_v$). The result $\tilde{t}_v^{(l)}$ is stored in memory as an updated representation of the node’s semantic environment.

    \item \textbf{Few-Shot Example Collection.}  
    From among the labeled structural neighbors, a small subset is selected as in-context examples. Each selected node contributes a \texttt{(text, label)} pair, which is explicitly written into the current node’s memory. This supports few-shot label prediction by providing semantically relevant examples during prompt construction:
    \[
    \mathcal{E}_v^{(l)} = \left\{ ( \tilde{t}_u^{(l-1)}, y_u ) \,\middle|\, u \in \mathcal{N}_{\text{local}}(v) \land y_u \in \mathcal{Y} \right\}
    \]
\end{itemize}

Thus, Local Aggregation provides both \textit{semantic enhancement} and \textit{label grounding} through neighboring information. All aggregated summaries and selected examples are written into the node’s memory buffer for use in future planning or prediction. The memory update is:
\[
\mathcal{M}_v^{(l)} \leftarrow \mathcal{M}_v^{(l-1)} \cup \{ \tilde{t}_v^{(l)} \} \cup \mathcal{E}_v^{(l)}
\]

\subsubsection{Global Aggregation.}

The \texttt{Global Aggregation} action enables a node to augment its semantic context by retrieving structurally distant but semantically similar nodes from across the graph. This is achieved by invoking a retrieval tool (e.g., RAG), which queries a structure-free database constructed from all node-level textual representations.
Unlike the original graph, this database preserves no edges; each entry consists of a node’s text feature $t_u$ and its label $y_u$ (if available). Retrieval is based purely on textual semantic similarity, typically computed via dense embedding similarity (e.g., cosine distance). Given a query node’s current textual state $t_v^{(l-1)}$, the system retrieves the Top-$K$ most relevant global semantic neighbors:
\[
\mathcal{N}_{\text{global}}(v) = \texttt{RAG}(t_v^{(l-1)}, top = K)
\]

Similar to Local Aggregation, Global Aggregation also serves two main purposes:

\begin{itemize}
    \item \textbf{Feature Enhancement.}  
    The textual features of the retrieved nodes are aggregated to form an updated summary $\tilde{t}_v^{(l)}$ that reflects global semantics, which is appended to the node’s memory to enhance its contextual understanding beyond structural proximity:
    \[
    \tilde{t}_v^{(l)} = \texttt{TextAgg} \left( \tilde{t}_v^{(l-1)}, \left\{ \tilde{t}_u^{(l-1)} \,\middle|\, u \in \mathcal{N}_{\text{global}}(v) \right\} \right)
    \]

    \item \textbf{Few-Shot Example Collection.}  
    From the retrieved nodes with known labels, a subset is selected as few-shot examples for prompt construction:
    \[
    \mathcal{E}_v^{(l)} = \left\{ ( \tilde{t}_u^{(l-1)}, y_u ) \,\middle|\, u \in \mathcal{N}_{\text{global}}(v) \land y_u \in \mathcal{Y} \right\}
    \]
\end{itemize}

Both the aggregated summary and the selected labeled examples are written into the node’s memory, which is the same as the local version.
Through this process, Global Aggregation provides semantic enrichment and label grounding from structure-agnostic sources. It expands each node’s informational horizon beyond its local neighborhood—especially benefiting nodes in sparse or isolated regions. This mechanism directly addresses \textit{Challenge 2 (Global Semantic Access)}.

\subsubsection{NoOp.}
While seemingly trivial, the \textbf{NoOp} (no operation) action plays a critical role in regulating information flow. When selected, the node intentionally chooses to take no action in the current layer—effectively pausing further aggregation or decision-making.

This mechanism is crucial for preventing information overload, particularly in situations where the memory already contains sufficient context. It helps prevent noise accumulation and supports pacing in multi-layer reasoning. By allowing nodes to wait or opt out of message passing altogether, NoOp reinforces ReaGAN’s core principle of self-decision and resource-aware adaptation (\textit{Challenge 1}).

\subsection{Memory: Tracking the Internal State}
\label{sec:memory}

As mentioned above, each node maintains a private memory buffer $\mathcal{M}_v$ that accumulates information over time to support reasoning and prediction~\citep{guo2026memeyevisualcentricevaluationframework}. This memory includes two types of content from two sources:
\begin{itemize}
    \item \textbf{Local information}: messages and labeled examples from Local Aggregation.
    \item \textbf{Global information}: semantically similar content retrieved via Global Aggregation.
\end{itemize}

\begin{figure*}[t]
    \centering
    \includegraphics[width=1.0\linewidth]{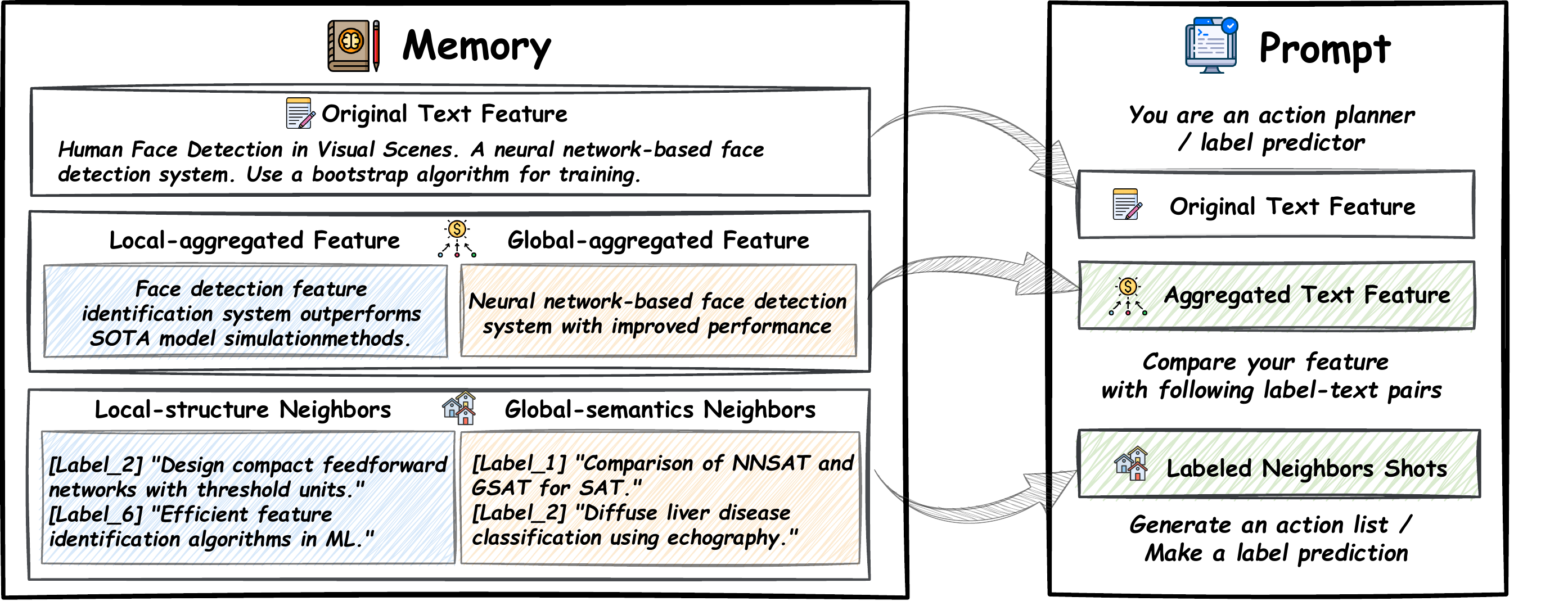}  
    \caption{
    \textbf{Information flow from memory to prompt.}
    Each node’s memory includes its original text feature, aggregated text feature from local and global neighbors, and selected labeled neighbor shots. 
    During planning or prediction, these components are selectively injected into a natural language prompt, providing the LLM with (i) the input node’s raw identity, (ii) context-enhanced descriptions, and (iii) label-text pairs for few-shot learning. 
    This design enables each agent to reason over multi-scale context and take personalized actions. 
    }
    \label{fig:memory_to_prompt}
\end{figure*}

As illustrated in Figure~\ref{fig:memory_to_prompt}, memory entries can also be categorized along a second semantic dimension: in addition to the \textit{source type} (local vs. global), we also consider the \textit{information purpose}. These correspond to the following categories:

\begin{itemize}
    \item \textbf{Text Feature.} The node’s raw natural language input $t_v$, preserved across all layers to serve as an identity anchor.

    \item \textbf{Aggregated Representations.} Natural language summaries collected via Local Aggregation and Global Aggregation. These capture multi-scale contextual signals to enrich node understanding.

    \item \textbf{Selected Labeled Neighbors.} A curated set of \texttt{(text, label)} examples drawn from both local and global sources. These are explicitly stored for use in few-shot prediction, injected into the prompt to support semantic reasoning.
\end{itemize}


Memory is updated incrementally at each layer by adding newly generated entries from the current step:
\[
    \mathcal{M}_v^{(l)} \leftarrow \mathcal{M}_v^{(l-1)} \cup \{ m_i^{(l)} \}_{i=1}^{k}
\]
where $\{ m_i^{(l)} \}$ are the new entries produced by actions such as Local Aggregation, Global Aggregation, or retrieval-based example selection at layer $l$, where the new entries consist of $\tilde{t}_v^{(l)}$ and $\mathcal{E}_v^{(l)}$.

As the core source of contextual information, the memory buffer provides the essential components for prompt construction: the original text, aggregated text, and a subset of labeled examples. This allows the prompt to accurately reflect the node’s internal state and accumulated knowledge. As each node independently controls its memory and its content evolves through executed actions, this design supports self-individualized behavior (\textit{Challenge 1}). Moreover, by storing semantically retrieved examples from distant nodes, memory also supports context enrichment beyond local structure (\textit{Challenge 2}).

\subsection{Tools: Global Semantic Augmentation}
\label{sec:tool}

To support global semantic reasoning, ReaGAN equips each node with a single external tool: Retrieval-Augmented Generation (RAG). This tool enables a node to retrieve semantically relevant information from across the entire graph—beyond its structural neighborhood. Because text embedding and retrieval quality can be affected by semantic shift, and system efficiency can further benefit from cache-aware agent-RAG designs, robust semantic retrieval is especially important in this setting~\citep{gao2026semanticshift, lin2025cache, zhu2026trustorabstainselfawareragapproach}.
A structure-free database is constructed, consisting of all nodes’ text features and, when available, their labels. Each node contributes a single entry, and the resulting corpus $\mathcal{D}$ is indexed by textual similarity. Given a node’s current representation $t_v^{(l)}$, the RAG tool performs a top-$K$ similarity search over the database:
\[
    \texttt{RAG}(t_v^{(l)}, top = K) = \texttt{TopK}\left( \left\{ t_u \,\middle|\, u \in \mathcal{V} \right\}, \texttt{sim}(t_v^{(l)}, t_u) \right)
\]
where $\texttt{sim}(\cdot,\cdot)$ denotes an embedding-based similarity function (e.g., cosine distance).

In this design, RAG functions as a modular tool that enables semantic retrieval for the \texttt{Global Aggregation} action. This separation ensures that memory evolution remains fully governed by agent actions, without implicit tool-side updates. By invoking RAG on demand, each node can enrich its local structural context with globally relevant, semantically aligned information—particularly beneficial for nodes situated in sparse or disconnected regions. As such, RAG plays a central role in addressing \textit{Challenge 2 (Global Semantic Access)}.

\subsection{Structured Action Space and Execution Control}

Unlike end-to-end LLM-based classification approaches, ReaGAN adopts a structured and discrete action space to regulate agent behavior. Each node-agent is restricted to a finite set of executable operations: \textit{LocalAggregation}, \textit{GlobalAggregation}, \textit{Prediction}, and \textit{NoOp}. This design ensures that LLM outputs are interpreted as high-level decisions rather than free-form textual predictions.

The use of a discrete action vocabulary serves two purposes. First, it constrains the reasoning space of the language model, preventing hallucinated or semantically invalid operations. Second, it enables a deterministic execution layer to mediate between LLM decisions and graph state transitions. Instead of allowing the LLM to directly manipulate node states, its output is parsed into structured actions, which are then validated and executed by a controlled runtime environment.

To enhance robustness, we introduce an action validation and fallback mechanism. When the LLM produces malformed or semantically inconsistent outputs, the system automatically reconstructs a legal action through a fallback procedure. Invalid actions are either converted into a well-formed \textit{Prediction} operation or downgraded to \textit{NoOp}, ensuring that execution never breaks due to generation errors. This design decouples reasoning from state mutation, significantly improving system stability in multi-step agent interactions.

Overall, the structured action interface transforms the LLM from a direct classifier into a decision-making policy component embedded within a regulated execution loop. This architectural separation is critical for maintaining controllability and reproducibility in decentralized graph reasoning.

\subsection{Memory System Design}

A core component of ReaGAN is its dual-memory architecture, which separates \textit{local memory} from \textit{global memory}. This distinction enables nodes to reason over neighborhood-level semantic signals while retaining the ability to access remote information when necessary.

\textbf{Local Memory.} Local memory is constructed through neighbor aggregation and broadcast operations. When a node executes \textit{LocalAggregation}, it retrieves structured memory entries from adjacent nodes. Broadcast operations further allow labeled training nodes to function as semantic anchors, propagating reliable information throughout the graph. To prevent uncontrolled memory growth, each memory entry follows a standardized schema containing text, label, source type, and layer information.

\textbf{Global Memory.} Global memory is accessed via retrieval-augmented generation (RAG). When local information is insufficient, a node may trigger \textit{GlobalAggregation}, querying a broader semantic pool. To improve retrieval quality, we incorporate a reranking mechanism with label diversity control, ensuring that selected examples are not overly concentrated on a single category.

\textbf{Memory Filtering and Deduplication.} To maintain stability across layers, we implement a memory deduplication mechanism based on structured entry comparison and normalized text matching. Duplicate entries are filtered using both structural attributes and semantic equivalence checks. Additionally, only the top-$K$ most relevant labeled examples are injected into the prompt context to prevent context overflow and reduce noise. This filtering strategy improves reasoning clarity while preserving representative supervision signals.

By combining localized propagation with controlled global retrieval, the memory system enables adaptive reasoning behaviors without relying on a centralized controller. The explicit memory management layer is essential for maintaining long-horizon consistency in decentralized agent interactions, and is closely related to recent work on memory sharing across LLM-based agents~\citep{gao2024memorysharinglargelanguage}.

\paragraph{Overall Execution Flow}

ReaGAN transforms each node into an autonomous agent equipped with memory, planning, and external tools. At each layer, nodes operate independently based on their internal state—without global synchronization or shared parameters.

Each reasoning layer follows the following cycle:
\begin{itemize}
    \item \textbf{Perception:} The node gathers contextual signals from its memory and optionally from local neighbors.
    \item \textbf{Planning:} It constructs a prompt and queries the frozen LLM to decide the next action(s).
    \item \textbf{Action:} The selected actions are executed—aggregating information, retrieving global content, or making a prediction—and the outcomes are written into memory.
\end{itemize}

This process repeats for $L$ layers (typically 1–3), after which each node outputs a predicted label. Through this decentralized execution mechanism, ReaGAN fulfills two key objectives:  
(1) enabling \textbf{node-level autonomy and personalized decision-making} (\textit{Challenge 1}); and  
(2) integrating \textbf{local structural and global semantic information} within a unified agentic framework (\textit{Challenge 2}).

\section{Experiments}

Based on the challenges identified in Section~\ref{sec:introduction}, we formulate the following research questions(RQs):

\begin{itemize}
    \item \textbf{RQ1}: How does ReaGAN perform on node classification tasks compared to standard GNNs?
    \item \textbf{RQ2}: How do the agentic planning mechanism and Global Aggregation contribute to performance?
    \item \textbf{RQ3}: When do agentic nodes need global semantic retrieval to do global aggregation, and how should local and global examples be balanced in prompts?
    \item \textbf{RQ4}: Does exposing label semantics improve classification accuracy in ReaGAN?
\end{itemize}

\begin{table}[!tb]
\centering
\caption{
Test accuracy (\%) on the Cora, Citeseer, and Chameleon datasets. Traditional GNNs rely on parametric training and fixed message passing, while \textbf{ReaGAN} leverages a frozen LLM for agentic planning and personalized reasoning.
}
\label{tab:main-results}
\begin{tabular}{lccc}
\toprule
\textbf{Model} & \textbf{Cora} & \textbf{Citeseer} & \textbf{Chameleon} \\
\midrule
\multicolumn{4}{l}{\textit{Parametric Training (Supervised GNNs)}} \\
\midrule
GCN~\citep{kipf2017semi}         & 84.71 & 72.56 & 28.18 \\
GAT~\citep{velivckovic2017graph} & 76.70 & 67.20 & 42.93 \\
GraphSAGE~\citep{hamilton2017inductive} & 84.35 & 78.24 & 62.15 \\
GPRGNN~\citep{chien2021adaptive} & 79.51 & 67.63 & 67.48 \\
APPNP~\citep{klicpera2018predict} & 79.41 & 68.59 & 51.91 \\
MLP-2~\citep{zhu2022revisiting}  & 76.44 & 76.25 & 46.72 \\
MixHop~\citep{abu2019mixhop}     & 65.65 & 49.52 & 36.28 \\
\midrule
\multicolumn{4}{l}{\textit{Frozen LLM + Few-shot Setting}} \\
\midrule
\textbf{ReaGAN (Ours)} & 84.95 $\pm$ 0.35 & 60.25 $\pm$ 0.36 & 43.80 $\pm$ 0.65 \\
\bottomrule
\end{tabular}
\end{table}

\paragraph{Datasets}
We evaluate the proposed ReaGAN on the standard node classification task using the \textbf{Cora} ,\textbf{Citeseer} and \textbf{Chameleon} dataset. Each node corresponds to a scientific publication with a textual description, and the goal is to predict its research category. We adopt a 60\%/20\%/20\% split for training, validation, and testing, respectively.

\paragraph{Baselines.}
We compare ReaGAN with a set of widely used baselines, including GCN~\citep{kipf2017semi}, GAT~\citep{velivckovic2017graph}, GraphSAGE~\citep{hamilton2017inductive}, APPNP~\citep{klicpera2018predict}, GPRGNN~\citep{chien2021adaptive}, MixHop~\citep{abu2019mixhop}, and MLP-2. All models are implemented using PyTorch Geometric with standard hyperparameters and are trained on the same data splits as ReaGAN.

\paragraph{Reproduction Settings.}
We conduct all experiments in PyTorch using a cluster of NVIDIA RTX A6000 GPUs. The agentic node reasoning module relies on a frozen LLM, served via vLLM, using Qwen2.5-14B-Instruct. We do not perform any fine-tuning. We report test accuracy for node classification. To ensure fair comparison, all baselines are trained under identical splits. We use all-MiniLM-L6-v2 to transfer text to embedding. More experimental details are provided in the Appendix. \ref{apx:hyper}.

\subsection{Overall Performance on Node Classification}
To answer \textbf{RQ1}, we compare ReaGAN's node classification accuracy against a suite of fully supervised GNN baselines, with the results detailed in Table~\ref{tab:main-results}. The findings demonstrate that ReaGAN, despite operating with a frozen LLM and without any gradient-based training, achieves remarkably competitive performance.
On the \textbf{Cora} dataset, ReaGAN achieves an accuracy of $84.95\%$, slightly outperforming well-established models like GCN ($84.71\%$) and GraphSAGE ($84.35\%$). This result is particularly noteworthy as it shows that an agentic, reasoning-based approach can match or even exceed the performance of models that have been explicitly trained on the graph's structure and features for this specific task.
On the \textbf{Citeseer} and \textbf{Chameleon} datasets, while some supervised methods like GraphSAGE and GPRGNN show higher accuracy, ReaGAN still maintains a strong competitive standing. For instance, its performance on Citeseer ($60.25\%$) is comparable to several established GNNs. It is crucial to frame this result in the context of our method's paradigm: ReaGAN is not fine-tuned on the downstream task. Instead, it leverages the pre-existing reasoning capabilities of an LLM to interpret the graph context in a zero-shot, few-shot manner.
In summary, these results validate the core premise of our work. By empowering nodes with autonomy and integrating both local and global information through a reasoning-based framework, ReaGAN establishes itself as a viable and powerful new paradigm for graph machine learning, one that trades task-specific training for greater flexibility and emergent reasoning capabilities.

\subsection{Agent Behavior Analysis}

Beyond quantitative accuracy, we analyze the behavioral patterns of node-agents across reasoning layers. Empirically, early layers tend to favor \textit{LocalAggregation}, relying primarily on neighborhood-level information to form initial beliefs. As reasoning progresses, the frequency of \textit{GlobalAggregation} increases, particularly for nodes located in sparse or ambiguous regions of the graph. This suggests that global semantic access functions as a compensatory mechanism when local signals are insufficient.

Prediction actions are typically concentrated in later layers, indicating that nodes accumulate sufficient contextual evidence before committing to a final label. In contrast, \textit{NoOp} actions are relatively rare and mainly occur in stable states where additional reasoning does not improve confidence.

We also observe that broadcast operations originating from labeled training nodes contribute to stabilizing local clusters. These nodes effectively serve as knowledge anchors, reducing prediction variance in nearby regions. Overall, the layered distribution of actions reflects a progressive reasoning process: local evidence gathering, optional global retrieval, and delayed commitment through prediction. This behavior-level analysis supports the claim that ReaGAN operates as a decentralized reasoning system rather than a single-step classifier.

\subsection{Impact of Agentic Planning and Global Retrievals}

To answer \textbf{RQ2}, we perform ablations on ReaGAN’s two core components: node-level planning and global retrieval. Removing prompt planning forces all nodes to follow a fixed action sequence, leading to notable accuracy drops due to the lack of context-specific behavior. Disabling global retrieval (Local Only) limits each node to structural neighbors, which underperforms in sparse graphs. Conversely, the Global Only variant removes structural input and struggles in well-connected graphs, showing the need for both structural and semantic signals. Experimental results are shown in Figure~\ref{fig:ablation}, which shows that both \textit{agentic planning} and \textit{global semantic access} are essential. Removing either component leads to a measurable degradation in accuracy, validating the effectiveness of our solution in addressing autonomy and semantic challenges.

\FloatBarrier
\begin{figure}[H]
\centering
\begin{tikzpicture}
\begin{axis}[
    width=\columnwidth,
    height=0.62\columnwidth,
    ybar,
    bar width=6.2pt,
    ymin=20,
    ymax=90,
    enlarge x limits=0.18,
    ylabel={Test Accuracy (\%)},
    symbolic x coords={Cora,Citeseer,Chameleon},
    xtick=data,
    x tick label style={font=\small},
    yticklabel style={font=\small},
    label style={font=\small},
    legend style={
        font=\scriptsize,
        at={(0.5,1.02)},
        anchor=south,
        legend columns=2,
        /tikz/every even column/.append style={column sep=6pt}
    },
    grid=both,
    grid style={line width=.1pt, draw=gray!20},
    major grid style={line width=.2pt,draw=gray!25},
]

\addplot+[draw=black, fill={rgb,255:red,140;green,197;blue,110}] coordinates {(Cora,84.95) (Citeseer,60.25) (Chameleon,43.80)};
\addlegendentry{ReaGAN (Full)}

\addplot+[draw=black, fill={rgb,255:red,198;green,224;blue,180}] coordinates {(Cora,79.83) (Citeseer,35.87) (Chameleon,38.29)};
\addlegendentry{No Planning}

\addplot+[draw=black, fill={rgb,255:red,170;green,210;blue,235}] coordinates {(Cora,81.67) (Citeseer,58.73) (Chameleon,25.60)};
\addlegendentry{Local Only}

\addplot+[draw=black, fill={rgb,255:red,215;green,230;blue,245}] coordinates {(Cora,79.67) (Citeseer,33.45) (Chameleon,24.94)};
\addlegendentry{Global Only}

\end{axis}
\end{tikzpicture}
\caption{
Ablation study on Cora, Citeseer, and Chameleon (Test Accuracy \%). 
ReaGAN’s performance depends on both agentic planning and local-global integration.
}
\label{fig:ablation}
\end{figure}

\begin{figure}[H]
\centering
\begin{tikzpicture}
\begin{axis}[
    width=\columnwidth,
    height=0.58\columnwidth,
    ybar,
    bar width=9.0pt,
    ymin=30,
    ymax=90,
    enlarge x limits=0.22,
    ylabel={Accuracy (\%)},
    symbolic x coords={Cora,Citeseer,Chameleon},
    xtick=data,
    x tick label style={font=\small},
    yticklabel style={font=\small},
    label style={font=\small},
    legend style={
        font=\scriptsize,
        at={(0.5,1.02)},
        anchor=south,
        legend columns=2,
        /tikz/every even column/.append style={column sep=6pt}
    },
    grid=both,
    grid style={line width=.1pt, draw=gray!20},
    major grid style={line width=.2pt,draw=gray!25},
]

\addplot+[draw=black, fill={rgb,255:red,170;green,210;blue,235}] coordinates {(Cora,84.95) (Citeseer,50.14) (Chameleon,43.80)};
\addlegendentry{Strategy A}

\addplot+[draw=black, fill={rgb,255:red,140;green,197;blue,110}] coordinates {(Cora,83.02) (Citeseer,60.25) (Chameleon,38.29)};
\addlegendentry{Strategy B}

\end{axis}
\end{tikzpicture}
\caption{
\textbf{Prompt Memory Strategy vs. Accuracy.}
Comparison of two prompt construction strategies across three datasets.
Strategy~A includes both local and global memory in the prompt.
Strategy~B includes global memory only when fewer than two local entries are available.
}
\label{fig:prompt-strategy}
\end{figure}
\FloatBarrier

To address \textbf{RQ3}, we investigate how the balance between local and global memory in the prompt affects node classification accuracy. We perform an ablation study comparing two prompt construction strategies: Strategy~A, which includes both local and global memory in the prompt, and Strategy~B, which includes global memory only when fewer than two local examples are available.
The results in Figure~\ref{fig:prompt-strategy}, particularly Citeseer's contrasting performance, highlight the critical need for an adaptive retrieval strategy~\citep{wang2026ragrouter, wei2026skillragfailurestateawareretrievalaugmentation, mei2025omnirouterbudgetperformancecontrollable}. The superior performance of Strategy~B on Citeseer is not arbitrary; it is a direct consequence of the underlying graph topology. A quantitative analysis reveals that Citeseer is a significantly sparser graph than Cora. Based on the dataset statistics in Table~\ref{tab:dataset-stats}, Citeseer has an average node degree of approximately 2.84 ($4,732 \text{ edges} / 3,327 \text{ nodes} \times 2$), which is considerably lower than Cora's average degree of roughly 4.01 ($5,429 \text{ edges} / 2,708 \text{ nodes} \times 2$).
This structural sparsity has a crucial implication: many nodes in Citeseer have small, less informative local neighborhoods. In such a setting, Strategy~A---which indiscriminately performs global retrieval---risks introducing ``semantic noise.'' While retrieved nodes may be textually similar, they might lack the precise contextual relevance that structural proximity provides, thereby confusing the agent and degrading performance.
In contrast, Strategy~B functions as a more intelligent and cautious mechanism. It prioritizes the high-fidelity signals from the immediate structural neighborhood and only resorts to global retrieval as a fallback when local information is critically insufficient (i.e., fewer than two local examples are available). This selective approach prevents the agent from diluting a sufficient local context with potentially noisy global information, leading to a more robust and accurate classification on sparser, ``noisier'' graphs like Citeseer.

Regarding \textbf{RQ4}, we investigate the impact of label semantics on the agent's reasoning process. The results, presented in Figure~\ref{fig:label-semantics}, are unequivocal: providing explicit, human-readable label names (e.g., "Machine Learning", "Databases") consistently degrades classification accuracy across all datasets.
This performance drop reveals a critical insight into the behavior of LLMs in graph-based reasoning tasks. When exposed to semantic label names, the LLM leverages its vast pre-trained knowledge to take a cognitive ``shortcut.'' Instead of reasoning from the evidence provided in the prompt---such as the node's features and the contextual examples from its neighbors---the model tends to overfit to the literal meaning of the labels. For instance, if a node's text contains keywords like ``algorithm'' or ``prediction,'' the LLM might impulsively select the ``Machine Learning'' label, even if the local and global examples in its memory point towards a more specific or different category. This creates biased guesses based on superficial keyword matching rather than a holistic analysis of the provided context.
Conversely, by anonymizing the labels (e.g., ``Label\_1'', ``Label\_2''), we compel the agent to perform a more genuine and robust reasoning task. Without the semantic shortcut, the LLM is forced to rely entirely on the in-context examples provided in the prompt to understand the defining characteristics of each category. It must compare the target node's features against the features of the labeled neighbors to infer the correct classification. This process ensures that the model's predictions are grounded in the specific, contextual evidence from the graph structure and node attributes, rather than its pre-existing and often overly general world knowledge. Our final design therefore adopts label anonymization as a crucial step to ensure faithful reasoning.

\FloatBarrier

\subsection{Further Analysis: System Efficiency and Agentic Behavior}
To further understand the internal mechanics and engineering robustness of ReaGAN, we profile the agent's action distribution, failure recovery rates, and inference costs.

\textbf{Adaptive Action Selection and Fallback Robustness.}
Log analysis reveals a significant behavioral shift as the system architecture matures, as visualized in Figure \ref{fig:action_shift}. In our naive early implementation, agents frequently struggled with sparse context, triggering expensive exploratory actions (e.g., \texttt{Broadcast} and \texttt{RAG Query}). In a sampled batch, a fallback mechanism was triggered 27 times, yet $100\%$ of these attempts failed to yield a valid state update, resulting in unrecoverable \texttt{None} predictions. By refining the agent's action space and memory formatting in our final design, the system exhibits remarkable robustness. The agent intelligently curtails noisy global queries in favor of high-confidence \texttt{Fallback Update} and local \texttt{Update} actions. In the optimized version, while fallback updates are still triggered under high uncertainty (33 times in a comparable batch), the execution success rate jumps to $100\%$, successfully mapping semantic reasoning to valid label IDs without a single \texttt{None} prediction.

\begin{figure}[htbp]
\centering
\begin{tikzpicture}
\begin{axis}[
    width=\columnwidth,
    height=0.58\columnwidth,
    ybar,
    bar width=10pt,
    ymin=35,
    ymax=90,
    enlarge x limits=0.32,
    ylabel={Accuracy (\%)},
    symbolic x coords={Cora,Citeseer},
    xtick=data,
    x tick label style={font=\small},
    yticklabel style={font=\small},
    label style={font=\small},
    legend style={
        font=\scriptsize,
        at={(0.5,1.02)},
        anchor=south,
        legend columns=2,
        /tikz/every even column/.append style={column sep=6pt}
    },
    grid=both,
    grid style={line width=.1pt, draw=gray!20},
    major grid style={line width=.2pt,draw=gray!25},
]

\addplot+[draw=black, fill={rgb,255:red,217;green,95;blue,2}] coordinates {(Cora,76.83) (Citeseer,42.11)};
\addlegendentry{Label Names Visible}

\addplot+[draw=black, fill={rgb,255:red,0;green,114;blue,178}] coordinates {(Cora,84.95) (Citeseer,60.25)};
\addlegendentry{Label\_ID only}

\node[font=\scriptsize, align=center] at (axis cs:Citeseer,87) {Sharp drop\\with semantics};
\draw[->, very thick] (axis cs:Citeseer,83) -- (axis cs:Citeseer,45);

\end{axis}
\end{tikzpicture}
\caption{
\textbf{Effect of Label Semantics on Accuracy.}
Showing label names (e.g., ``Rule Learning'') in the prompt harms performance,
as LLMs tend to overfit to label wording rather than reasoning from memory.
We anonymize all labels (e.g., ``Label\_2'') in our final design; label names are only revealed here for this controlled comparison.
}
\label{fig:label-semantics}
\end{figure}

\begin{figure}[htbp]
\centering
\begin{tikzpicture}
\begin{axis}[
    width=\columnwidth,
    height=0.58\columnwidth,
    ybar stacked,
    bar width=18pt,
    ymin=0,
    enlarge x limits=0.35,
    ylabel={Action Count},
    symbolic x coords={Early Version,Optimized Version},
    xtick=data,
    x tick label style={font=\small, align=center},
    yticklabel style={font=\small},
    label style={font=\small},
    legend style={
        font=\scriptsize,
        at={(0.5,1.02)},
        anchor=south,
        legend columns=2,
        /tikz/every even column/.append style={column sep=6pt}
    },
    grid=both,
    grid style={line width=.1pt, draw=gray!20},
    major grid style={line width=.2pt,draw=gray!25},
]

\addplot+[draw=black, fill=black!25] coordinates {(Early Version,0) (Optimized Version,33)};
\addlegendentry{Fallback Update}

\addplot+[draw=black, fill={rgb,255:red,0;green,114;blue,178}] coordinates {(Early Version,5) (Optimized Version,12)};
\addlegendentry{Update (Local)}

\addplot+[draw=black, fill={rgb,255:red,140;green,197;blue,110}] coordinates {(Early Version,19) (Optimized Version,3)};
\addlegendentry{Retrieve (Neighbor)}

\addplot+[draw=black, fill={rgb,255:red,217;green,95;blue,2}] coordinates {(Early Version,6) (Optimized Version,0)};
\addlegendentry{Broadcast}

\addplot+[draw=black, fill={rgb,255:red,231;green,138;blue,195}] coordinates {(Early Version,3) (Optimized Version,0)};
\addlegendentry{RAG Query (Global)}

\end{axis}
\end{tikzpicture}
\caption{Shift in agent action distribution. The optimized version autonomously shifts from expensive exploration (Broadcast, RAG Query) to robust local and fallback actions.}
\label{fig:action_shift}
\end{figure}

\textbf{Posterior Label Bias in Fallback Scenarios.}
When nodes are entirely isolated and agents are forced to rely on fallback decisions, we observe a distinct posterior distribution in the LLM's predictions. Out of 45 recorded fallback mapping events on the Cora dataset, the agent favored "Theory" (13 times), "Probabilistic Methods" (9 times), and "Neural Networks" (8 times), while rarely guessing "Rule Learning" (2 times). This skewed distribution indicates that in the absence of local structural evidence or global RAG context, the frozen LLM defaults to its pre-trained semantic priors. This underscores the necessity of our RAG and local aggregation modules---without them, the agent reverts to biased, keyword-driven guessing rather than graph-aware reasoning.

\textbf{Inference Efficiency and Scalability.}
A common critique of LLM-agentic systems is their inference latency. We evaluate ReaGAN's computational overhead to demonstrate its viability. Operating on an NVIDIA RTX A6000 GPU cluster with a 14B parameter frozen LLM (served via vLLM), the agentic reasoning process demonstrates stable, linear scaling. Processing a batch of 64 nodes requires approximately 3 minutes and 18 seconds, yielding a consistent inference cost of roughly $3.10$ seconds per iteration per batch. When scaling down to a 20-node batch, the time amortizes proportionally (maintaining $\sim 3.06$ to $3.22$ seconds per iteration). While this inference latency is higher than a forward pass of a simple MLP, it fundamentally bypasses the massive memory and time overhead associated with gradient-based fine-tuning on large graphs. Thus, ReaGAN offers a highly scalable, zero-gradient deployment alternative for complex graph reasoning.

\FloatBarrier

\section{Related Work}

\subsection{Graph Neural Networks and Message Passing}
Graph Neural Networks (GNNs) such as GCN{}~\citep{kipf2017semi}, GraphSAGE{}~\citep{hamilton2017inductive}, and GAT{}~\citep{velivckovic2017graph} have become the dominant paradigm in graph machine learning. These models rely on fixed, layer-wise message passing schemes that aggregate information from each node’s neighbors using predefined update functions. While effective, this rigid design limits expressiveness and can cause issues like over-smoothing and over-squashing.
To overcome these limitations, several works have proposed more flexible message passing strategies. CoGNN{}~\citep{finkelshtein2023cognn} introduces cooperative agents that decide whether to broadcast or listen during each round, enabling more adaptive communication. However, these agents still rely on handcrafted utility functions and rule-based execution.
In contrast, our method allows each node to independently plan and execute its own message passing actions using an LLM-based agentic mechanism, and extends the communication space beyond local structural neighbors to include global semantic neighbors.

\subsection{Large Language Models for Graph Tasks}
Recent work has explored the use of LLMs for graph learning{}~\citep{shu2024knowledge, zhao2024dynllmlargelanguagemodels}. PromptGFM{}~\citep{zhu2024llm} converts nodes into text-based prompts and uses LLMs to learn a graph vocabulary for classification. In-context RAG{}~\citep{li2024incontext} frames node classification as a retrieval-augmented generation task, where textual neighbors are fetched and provided to the LLM for reasoning. These methods effectively leverage LLM capabilities but do not enable node-level autonomy or action planning.
STGP{}~\citep{hu2024promptgraph}, a prompt-based transfer learning framework for spatio-temporal graphs, to tackle the challenge of cross-domain and cross-task generalization. By using a unified task template and a two-stage prompting mechanism, the framework allows a model to effectively adapt to multiple tasks in new, data-scarce domains, achieving significant performance improvements on tasks such as forecasting and kriging.
By contrast, we use LLMs not as passive inference engines but as active planners—the cognitive core of each agent node. The LLM decides what action to take next, what context to gather, and when to predict.

\subsection{Agent-Based Graph Learning}
Several prior studies have introduced the notion of “agents” in graphs, but differ from our formulation. AgentNet{}~\citep{martinkus2023agent} trains neural agents to walk the graph and distinguish structures via learned exploration policies, but these agents are part of a supervised model and do not make autonomous decisions. GAgN{}~\citep{liu2024gagn} also proposes node-agent architectures for adversarial defense and resilience, but use hardcoded 1-hop views or restricted inference logic rather than learned behaviors.
Our formulation is fundamentally different: we treat each node as a full-fledged intelligent agent, with the ability to observe, reason, and act using LLM-powered prompts, without predefined roles, hardcoded transitions, or limited interaction space.

\section{Conclusion}

We introduced ReaGAN, a novel framework that treats each node in a graph as an autonomous agent capable of planning, acting, and reasoning through interactions with a frozen LLM. Unlike traditional GNNs that apply fixed, synchronous message passing rules to all nodes, ReaGAN enables self-decision at the node level—empowering each node to determine what information to gather, how to interact with others, and when to make predictions based on its own memory and context. Through this agentic formulation, ReaGAN seamlessly integrates both local structural signals and global semantic information via a unified prompting interface. Each node operates independently, combining Local Aggregation, global retrieval, and memory-guided few-shot reasoning to support fully individualized behavior. 
Importantly, ReaGAN achieves competitive performance using only a frozen LLM, without any gradient-based training or model fine-tuning. This highlights the promise of structured prompting and autonomous planning as a plug-and-play alternative to traditional GNNs.
By shifting from rigid, fixed-rule aggregation to retrieval-augmented, node-specific decision making, ReaGAN opens a new direction for graph learning with LLM-powered agents.

\section{Limitations and Future Work}

ReaGAN views each node as an autonomous agent with perception, planning, memory, and action capabilities. Although the results are promising, the current framework has several limitations. First, invoking a frozen LLM over multiple reasoning layers incurs substantial inference latency and memory overhead relative to conventional GNNs. Future work could improve efficiency through model compression, caching, selective retrieval, and resource-aware routing or scheduling~\citep{mei2025omnirouterbudgetperformancecontrollable, mei2025aiosllmagentoperating}.

Second, the current evaluation focuses on node classification in static text-attributed graphs. Broader evaluation across additional LLM backbones, dynamic or heterogeneous graphs, and graph tasks such as link prediction, recommendation, and graph-level reasoning is needed to establish the generality of the approach~\citep{10.1007/978-3-031-30675-4_3}. The performance of the agent also depends on prompt construction, retrieved examples, and memory selection; systematic prompt optimization, learned memory editing, and more robust retrieval control remain important directions for improving reliability.

Finally, the node-agent formulation naturally extends beyond isolated graph classification. We envision ReaGAN as a foundational component for multi-agent systems~\citep{li2025knowropesheuristicstrategy, chen2025graph2evalautomaticmultimodaltask}, in which node agents coordinate through decentralized decision-making~\citep{huang2025desocialblockchainbaseddecentralizedsocial} and inter-agent communication. Its ability to combine local reasoning with global retrieval may also support modular or routing-based architectures~\citep{gershon2025infrastructurepoweringibmsgen, mei2025omnirouterbudgetperformancecontrollable}. These extensions could turn ReaGAN into a broader blueprint for scalable, context-aware, and communication-capable agentic inference.


\bibliographystyle{ACM-Reference-Format}
\bibliography{sample-base}

\newpage
\appendix

\section{Technical Appendices and Supplementary Material}

\subsection{Prompt Example}

\subsubsection{Planning Prompt Example}  

You are a node in a text-attributed graph. Your goal is to plan the next action(s) for yourself based on your current context, including memory, text features, and neighbors.
\\

\noindent You may choose one of the following actions:

- \texttt{local aggregate}: aggregate structure neighbors’ text features and store the labeled nodes in memory.

- \texttt{global aggregate}: retrieve semantically similar nodes from the whole graph and aggregate the text.

- \texttt{local+global aggregate}: perform both local and global aggregation in this step.

- \texttt{no\_op}: do nothing and move to the next layer.

\bigskip

\noindent Node State:  \\
- \texttt{Text Feature:} ``Adverse interaction with tree depth restriction.'' \\
- \texttt{Last Local Aggregated Text Feature:} ``Adverse with tree depth restriction in genetic programming.'' \\
- \texttt{Last Global Aggregated Text Feature:} ``Genetic machine learning algorithms in scheduling performance problem.'' \\
- \texttt{Memory:} Contains 4 labeled examples. \\

\bigskip

\noindent Labeled Examples in Memory:  

\noindent Local:

- \texttt{Label 1}: ``Genetic algorithms for various scheduling problems.''

\noindent Global:

- \texttt{Label 1}: ``Team dynamics and performance enhancement strategies.'' 

- \texttt{Label 2}: ``Diverse machine learning techniques approach.'' 

- \texttt{Label 6}: ``Efficient learning of rectangle unions.''

\bigskip

\noindent Planning Your Steps:

- Think like a planner: Your goal is to gather enough information for the final label prediction

- If you cannot predict the label yet(need more context to do prediction), please choose local aggregate or global aggregate.

- If local memory is not enough, do local aggregation; meanwhile, if global memory is not enough, do global aggregation. Their amount is better to be in balance.

- Otherwise, choose "no\_op". \\

\noindent Respond strictly in JSON:
\begin{verbatim}
[
  {"action_type": "local aggregate", 
   "global aggregate" or "no_op"},
  {"action_type": "local aggregate", 
    "global aggregate" or "no_op"}
]
\end{verbatim}

\subsubsection{Prediction Prompt Example}  

You are a label prediction agent. You will be given a new node's aggregated text feature along with a memory of labeled examples.  
Use the memory to infer the most likely label for this node.  
Respond strictly in the required JSON format.
\\

\noindent \textbf{Node State:} 

- \texttt{Text Feature:} ``Adverse interaction with tree depth restriction.'' 

- \texttt{Local Aggregated Text Feature:} ``Adverse with tree depth restriction in genetic programming.'' 

- \texttt{Global Aggregated Text Feature:} ``Genetic machine learning algorithms in scheduling performance problem.'' 

\bigskip

\noindent \textbf{Labeled Examples in Memory:}

\noindent Local:

- \texttt{Label 1}: ``Genetic algorithms for various scheduling problems.'' \\

\noindent Global:

- \texttt{Label 1}: ``Team dynamics and performance enhancement strategies.'' 

- \texttt{Label 2}: ``Diverse machine learning techniques approach.'' 

- \texttt{Label 6}: ``Efficient learning of rectangle unions.''

\bigskip

Your task is to choose the most appropriate label from the following candidates:

\texttt{["Label 0", "Label 1", "Label 2", "Label 3", "Label 4", "Label 5", "Label 6"]}

\noindent Please follow these steps in your analysis:

1. Analyze the Current Node Text:   
- Identify primary topics and application domain   
- Determine the specific problem being solved   
- Note core methodologies and algorithms" 

2. Analyze Memory Examples:   
- Understand application domains for each label   
- Identify types of problems addressed  
- Note underlying methodologies 

3. Compare and Weigh Evidence:   
- Prioritize domain and problem alignment
- Evaluate methodological congruence
- Consider both domain-specific techniques and general paradigms   
- Ensure holistic coherence in your decision 

4. Avoid over-reliance on isolated keywords \\

\noindent Please think step by step: 

First, analyze memory examples and their labels, then compare them to the input text. Identify the most semantically similar memory items and explain why. Finally, decide which label best matches and explain your reasoning.

\bigskip

\noindent Respond strictly in JSON:
\begin{verbatim}
{"action_type": "predict", 
 "predicted_label": "Label ID"}
\end{verbatim}

\subsection{Dataset Construction and Processing}
\label{apx:dataset}

We evaluate ReaGAN on three benchmark datasets: Cora, Citeseer, and Chameleon. Cora and Citeseer are standard citation networks, where nodes represent scientific papers and edges indicate citation links. Node labels correspond to research topics. Chameleon is a Wikipedia-based graph with topic-labeled web pages as nodes and hyperlink edges. Other information of these datasets are shown in Table~\ref{tab:dataset-stats}

\begin{table}[ht]
\centering
\caption{Dataset statistics for node classification benchmarks.}
\label{tab:dataset-stats}
\begin{tabular}{lccc}
\toprule
\textbf{Dataset} & \textbf{\# Nodes} & \textbf{\# Edges} & \textbf{\# Classes} \\
\midrule
Cora       & 2,708  & 5,429   & 7 \\
Citeseer   & 3,327  & 4,732   & 6 \\
Chameleon  & 2,277  & 36,101  & 5 \\
\bottomrule
\end{tabular}
\end{table}

For each node, we extract a natural language text input. In Cora and Citeseer, this is obtained by concatenating the paper's title and abstract. In Chameleon, we use the full raw text provided in the dataset. To reduce prompt length and maintain LLM efficiency, we optionally apply a frozen LLM to compress these texts before use.

All node texts are stored in plain natural language format and are used in both planning and prediction prompts.

\subsection{Hyperparameter Settings}
\label{apx:hyper}

\paragraph{ReaGAN Settings}

\begin{itemize}
    \item LLM: Qwen2.5-14B-Instruct (frozen; served via vLLM backend)
    \item Prompt length: 512 tokens
    \item RAG Top-K: 5
    \item Max reasoning layers: 3 
    \item Few-shot examples per node: up to 5 from local neighbors and 5 from global retrieval
    \item Label prediction is enforced after the final layer
\end{itemize}







\end{document}